\documentclass[final,3p]{elsarticle}

\usepackage{subfigure,gensymb,graphicx,booktabs}
\usepackage[abs]{overpic}
\usepackage{textcomp,latexsym,tikz,array}
\usepackage{booktabs,threeparttable}
\usepackage[colorlinks]{hyperref}
\usepackage[ruled,vlined,linesnumbered]{algorithm2e}
\usetikzlibrary{matrix}
\journal{BioSystems}
\biboptions{authoryear}

\makeatletter
\def\ps@pprintTitle{%
 \let\@oddhead\@empty
 \let\@evenhead\@empty
 \def\@oddfoot{\small{BioSystems~\url{https://doi.org/10.1016/j.biosystems.2019.05.005}}}%
 \let\@evenfoot\@oddfoot}
\makeatother

\begin{document}

\begin{frontmatter}

\title{Towards an evolvable cancer treatment simulator}

\author[csct]{Richard~J.~Preen\corref{correspondingauthor}}
\ead{richard2.preen@uwe.ac.uk}
\cortext[correspondingauthor]{Corresponding author}
\address[csct]{Department of Computer Science and Creative Technologies}
\address[ucg]{Unconventional Computing Laboratory \\ University of the West of England, Bristol, BS16 1QY, UK}
\author[csct]{Larry~Bull}
\author[ucg]{Andrew~Adamatzky}

\begin{abstract}
The use of high-fidelity computational simulations promises to enable high-throughput hypothesis testing and optimisation of cancer therapies. However, increasing realism comes at the cost of increasing computational requirements. This article explores the use of surrogate-assisted evolutionary algorithms to optimise the targeted delivery of a therapeutic compound to cancerous tumour cells with the multicellular simulator, PhysiCell. The use of both Gaussian process models and multi-layer perceptron neural network surrogate models are investigated. We find that evolutionary algorithms are able to effectively explore the parameter space of biophysical properties within the agent-based simulations, minimising the resulting number of cancerous cells after a period of simulated treatment. Both model-assisted algorithms are found to outperform a standard evolutionary algorithm, demonstrating their ability to perform a more effective search within the very small evaluation budget. This represents the first use of efficient evolutionary algorithms within a high-throughput multicellular computing approach to find therapeutic design optima that maximise tumour regression.
\end{abstract}


\end{frontmatter}

%\linenumbers

\section{Introduction}

% intro to PhysiCell
PhysiCell~\citep{Ghaffarizadeh:2018} is an open source\footnote{PhysiCell source code (BSD license): \url{https://github.com/MathCancer/PhysiCell}} multicellular simulator based on the biotransport solver, BioFVM~\citep{Ghaffarizadeh:2016}. BioFVM simulates substrate secretion, diffusion, uptake, and decay; while PhysiCell models the emergent tissue-scale behaviour by simulating cell cycling, death, volume, mechanics, and motility. PhysiCell enables the simulation of new environmental substrates, cell types, and systems of cells, providing a general-purpose toolkit for exploring multicellular systems.

% intro to anti-cancer simulation
\cite{Norton:2019} present a review of agent-based models used to study cancer-immune interaction and immunotherapy; and \cite{Metzcar:2019} provide a more general overview of cell-based computational modelling in cancer biology. Many proposed cancer therapies attempt to target malignant cells by finding specific surfaces or molecules in order that drugs can be conjugated to custom antibodies or encapsulated in custom nanoparticles. \cite{Ghaffarizadeh:2018} introduced the agent-based PhysiCell 2-D anti-cancer biorobots simulation. This began the design of cell-cell interaction rules to create a multicellular cargo delivery system that actively delivers a cancer therapeutic beyond regular drug transport limits to hypoxic cancer regions. These model rules were manually tuned to achieve this (as yet unoptimised) design objective, requiring weeks of people-hours to configure, code, test, visualise, and evaluate~\citep{Ozik:2018a}.

\cite{Ghaffarizadeh:2018} also presented 3-D simulations of cancer immunotherapy. Using this simulator, \cite{Ozik:2018a} performed a human-selected parameter sweep (27 parameter sets; each set sampled 10 times) with each simulation requiring $\approx2$ days to complete. The results provided insights into therapeutic failure, thus demonstrating the potential of high-throughput computing to investigate high dimensional cancer simulator parameter spaces. High-throughput model investigation and hypothesis testing provides a new paradigm for solving complex problems, gaining new insights, and improving cancer treatment strategies~\citep{Ozik:2018a}.

% overview
Surrogate model-based optimisation has long been used in applications requiring expensive parameter evaluations, whether via simulated or physical testing~\citep{Sacks:1989,Jones:1998,Settles:2010}. In this article, we explore the use of surrogate-assisted evolutionary algorithms (EAs) to sequentially optimise the targeted delivery of a therapeutic compound to cancerous tumour cells with the multicellular simulator, PhysiCell. This represents the first use of efficient EAs within a high-throughput multicellular computing approach to find therapeutic design optima that maximise tumour regression.

\section{Background}

% intro to surrogate modelling
The use of surrogate models to reduce the number of costly EA fitness evaluations can be traced back to the origins of the discipline~\citep{Dunham:1963}. \cite{Jin:2018} present an overview of the process: first the variables to optimise are chosen; some initial parameters are then evaluated; a surrogate (regression) model is selected and used to build a model of the evaluated parameters; followed by a search of the model to identify new parameter values to evaluate; and the newly evaluated parameters added to the existing data. The process loops via continued iterations of model building, searching, and evaluation of selected parameters.

% selecting surrogate model
Many regression models have been used as a surrogate for the real fitness function, including multi-layer perceptron (MLP) based artificial neural networks~\citep[e.g.,][]{Bull:1997}, Gaussian processes~\citep[GP; also known as Kriging; e.g.,][]{Liu:2014}, radial basis functions~\citep[RBF; e.g.,][]{Regis:2014}, support vector regression~\citep[e.g.,][]{Yun:2009}, particle swarm optimisation~\citep[e.g.,][]{Wang:2017}, Markov networks~\citep[e.g.,][]{Brownlee:2013}, and coevolved fitness predictors~\citep{Schmidt:2008}. Many ensembles have also been explored~\citep[e.g.,][]{Wang:2018}. Probabilistic models such as GP are perhaps the most widely used surrogate since they provide a measure of confidence that can be used to efficiently select samples for evaluation~\citep{Jin:2018}. \cite{Preen:2016} showed that with very small and noisy samples there is little difference between the modelling approaches, with MLPs appearing to be a robust approach capable of capturing the underlying structure of the search space.

% selection and size of training data
The computational complexity of model building relative to the sampling expense is a key consideration, particularly in the case where large archive sets are used for training; for example, the computation time for GP training increases in cubic with the number of training data~\citep{Shahriari:2016}. MLP models are typically adopted when there are a large number of decision variables and/or training data~\citep{Chugh:2017}. This leads to a further key consideration, which is whether to use the full archive set for model training. It may be necessary to restrict the number of samples for use with a computationally complex model such as GP when there is a large archive set~\citep{Jin:2018}. Additionally, if there are any significant temporal affects, issues surround how best to select the subset for training~\citep{Preen:2017}.

% selecting samples to evaluate
The task of selecting which samples to evaluate is analogous with the problem of active learning wherein the algorithm is able to interactively query an oracle (e.g., a user) to obtain the output at a given data point~\citep{Settles:2010}. For online data-driven optimisation, initial data collection is often simply performed via random selection or a design of experiments technique such as Latin hypercube sampling~\citep{Jin:2018}. Subsequent samples are selected through the use of an acquisition function (also known as the infill sampling criteria) which rates the expected utility of evaluating a candidate solution~\citep{Shahriari:2016}. The most commonly used acquisition functions are the mean of model prediction, and the expected improvement~\citep[EI;][]{Jiao:2019}. Acquisition functions typically aggregate the model predicted fitness (e.g., mean) and estimated confidence (e.g., standard deviation) to explore regions of both high promise and high uncertainty. In addition to variance-based sampling, query-by-committee, cross-validation, and gradient-based methods also exist~\citep{Liu:2018}.

% model management
The use of the acquisition function is performed within an overall model management. \cite{Jin:2011} categorised approaches to model management as generational, individual, or population-based, depending upon whether whole generations, samples of individuals within a generation, or sub-populations are evaluated. The pre-selection approach~\citep{Emmerich:2002} uses the model to rate $M$ number of offspring and the best of these are chosen for evaluation on the real fitness function. As highlighted by \cite{Jin:2011}, the main difference between individual-based strategies and pre-selection is that the real fitness value is always used for selection, whereas in individual-based methods selection may be based on fitness values from the surrogate. This potentially makes the approach more robust to noise. Pre-selection has previously been shown effective for the evolution of (noisy) physical systems in which each design is physically instantiated and evaluated for fitness~\citep{Preen:2015,Preen:2016,Preen:2017}.

% optimisation uncertainty
Four main categories of uncertainty (or noise) are frequently encountered in real-world optimisation problems~\citep{Jin:2005}:
\begin{enumerate}
	\item Uncertainty in the objective function evaluation. This is common in stochastic simulations and physical systems where evaluations with the same parameters produce different results, e.g., by sensory measurement errors. Efficiently reducing the objective function error requires optimally selecting how many samples to perform for a given parameter set. Issues such as how best to aggregate the results, e.g., by mean or median may also affect convergence.
	\item The objective function is approximated by a surrogate model. Efficiently reducing the model error requires optimally selecting which samples to evaluate with the real objective function to provide new training data and avoidance of over-fitting through validation~\citep{Bischl:2012}. Approximated objective values are also encountered where the function evaluations take place over time and partial fitness scores can be observed by early termination~\citep{Park:2013}. Approximation error can be beneficial to the search process by aiding the escape of local optima~\citep{Lim:2010}.
	\item Dynamic optimisation wherein the objectives change over time~\citep[e.g.,][]{Chen:2018}. In these scenarios, population diversity and memory mechanisms are essential.
	\item Robust optimisation~\citep[e.g.,][]{Yu:2017} which seeks solutions that are less sensitive to small parameter perturbations, e.g., due to manufacturing tolerances. This typically involves a trade-off between solution quality and robustness.
\end{enumerate}

\cite{Rakshit:2017} present an overview on techniques for dealing with noisy evolutionary optimisation. Uncertainty in the objective function can be addressed via the explicit averaging of resampled parameters or implicit averaging using a large population size. The evolutionary selection mechanism may also be modified to account for noisy evaluations; for example, by only accepting offspring with observed fitness greater than the parent's plus some threshold. Repeated sampling is typically more effective for noise handling than parent populations and threshold selection, resulting in an exponential speed-up for noisy evolutionary optimisation in some cases~\citep{Qian:2018}. The number of samples to perform for a given parameter set can either remain fixed (static) for all candidate solutions or be dynamically allocated to each, e.g., based on the sample variance~\citep{Siegmund:2013}.

\section{Physics-based Multicellular Simulations}

\subsection{Methodology}
 
Here we use the PhysiCell 2-D anti-cancer biorobots simulator~\citep{Ghaffarizadeh:2018}. This performs a multicellular simulation of targeted drug delivery by modelling three types of cells:
\begin{itemize}
	\item Oxygen consuming cancer cells that generate a chemoattractant $c_1$ forming an oxygen gradient which can be used to guide worker cells.
	\item Worker cells that may adhere to cargo cells. Each cell has a persistence time, migration speed, migration direction, and migration bias. Worker cells perform a biased random migration towards cancer cells when adhered to cargo, and a biased random migration towards cargo cells when unadhered. Migration biases range [0,1] with 0 representing Brownian motion and 1 deterministic motion. The motility of unadhered worker cells is disabled if $c_1$ falls below a threshold.
	\item Cargo cells that secrete a diffusible chemoattractant $c_2$ used to guide worker cells. Adhered cargo cells detach from worker cells and secrete a therapeutic compound that induces apoptosis in nearby tumour cells when oxygenation falls below the cargo release o$_2$ threshold.
\end{itemize}

Each simulation is initialised with a 200~micron radius tumour. After 7 simulated days of tumour growth, 500 cells are ``injected'' near the tumour: 10\% worker cells and 90\% cargo cells. The simulation subsequently continues for 3 additional days of cancer therapy. A single simulation requires $\approx5$ minutes of wall-clock time on an Intel\textsuperscript{\textregistered} Xeon\textsuperscript{\textregistered} CPU E5-2650 v4 @ 2.20GHz with 64GB RAM using half of the 48 cores. Here we search the $N=6$ parameters specifying worker agent characteristics and cargo properties with the goal of minimising the resulting number of cancer cells after a period of simulated treatment. All other parameters remain at their original values as shown in Table~\ref{table:params}. 

\begin{table}[tbh]
	\caption{Default multicellular simulation parameters.}
    \centering
    \begin{tabular}{l l}
        \hline                
			Maximum attachment distance & 18~microns\\ 
			Minimum attachment distance & 14~microns\\ 
			Worker apoptosis rate & 0~minutes$^{-1}$\\ 
			Worker migration speed & 2~microns/minute\\ 
			Worker o$_2$ relative uptake & 0.1~minutes$^{-1}$\\ 
			Cargo o$_2$ relative uptake & 0.1~minutes$^{-1}$\\ 
			Cargo apoptosis rate & 4.065e-5~minutes$^{-1}$\\ 
			Cargo relative adhesion & 0\\
			Cargo relative repulsion & 5\\
			Damage rate & 0.03333~minutes$^{-1}$\\ 
			Repair rate & 0.004167~minutes$^{-1}$\\ 
			Drug death rate & 0.004167~minutes$^{-1}$\\ 
			Maximum relative cell adhesion distance & 1.25\\ 
			Elastic coefficient & 0.05~minutes$^{-1}$\\ 
			Maximum elastic displacement & 50~microns\\ 
			Motility shutdown detection threshold & 0.001\\
			Attachment receptor threshold & 0.1\\
        \hline
	\end{tabular}
	\label{table:params}
\end{table}

In particular, we explore the attached worker migration bias [0,1]; the unattached worker migration bias [0,1]; worker relative adhesion [0,10]; worker relative repulsion [0,10]; worker motility persistence time (minutes) [0,10]; and the cargo release o$_2$ threshold (mmHg) [0,20]. We adopt the surrogate-assisted pre-selection approach (and parameter values) previously used successfully to perform physical test-driven optimisation~\citep{Preen:2017}. As benchmark, a steady-state genetic algorithm (GA) with population size $P=20$ is used; tournament size $T=3$ for both selection and replacement; uniform crossover is performed with $\mathcal{X}=80\%$ probability; and a per allele mutation rate $\mu=1/N$ with a uniform random step size $s=[-5,5]\%$. A static sampling approach is used wherein $k$ simulation runs are performed for each candidate solution before assigning the fitness as the mean number of remaining cells after 7 simulated days of tumour growth plus 3 days of targeted drug delivery.

For the surrogate-assisted GA, all individuals in the initial population are evaluated and a regression model fit. Subsequently, evolution proceeds by iteratively selecting 2 parents via tournament and then creating and evaluating $M=1000$ offspring with the model via an acquisition function. The most promising of these is then selected for evaluation by the multicellular simulator and replaces an individual in the population selected via a negative tournament. Finally, the evaluated archive set is updated and the model retrained. The full archive set is used for training since there are no temporal affects on sampling, the time required to fit the model is insignificant in comparison with the sample evaluation time, and only a few decision variables are optimised. Algorithm~\ref{alg:sea} provides an outline of the surrogate-assisted GA with static sampling.

Since MLPs are extrapolative and GPs are interpolative, here we employ both approaches to model building and observe their effects on the evolved biophysical parameters. Here, the GP model~\citep{Rasmussen:2006} uses an RBF kernel function and the MLP uses $H=10$ rectified linear units in the hidden layer. Both models are trained using the limited-memory Broyden-Fletcher-Goldfarb-Shanno (L-BFGS) quasi-Newton optimisation algorithm~\citep{Byrd:1995}. As the GP model provides a measure of confidence (i.e., standard deviation) for each prediction, the EI is used as the acquisition (rating) function to select the next offspring; whereas the MLP does not provide a measure of confidence and so the best predicted fitness is used in this case. Algorithm~\ref{alg:ei} shows the EI rating function used for the GP model. For the MLP model, $Rating()$ in Algorithm~\ref{alg:sea} returns the model predicted fitness. All experiments are initialised with the same randomly generated population.

\begin{figure}[tbh]
    \begin{algorithm}[H]
    	\SetNoFillComment
    	\small
    	\DontPrintSemicolon
		$N=6$, $P=20$, $k=10$, $M=1000$, $\mathcal{X}=0.8$, $\mu=1/N$, $s=0.1$, $T=3$\;
		Initialise population $Pop = \{\vec{x}_1, \ldots, \vec{x}_P \}$ \tcp*{$\vec{x}$ normalised [-1,1]}
		Evaluate $Pop$ with the real objective function $k$ times and add to archive $\mathcal{A}$\;
		\While{evaluation budget not exhausted}{
			\tcc{build surrogate model}
			Fit regression model $\mathcal{R}$ using $\mathcal{A}$\;
			\tcc{pre-select offspring}
			Parent $p_1 \leftarrow TournamentSelection(Pop,T)$\;
			Parent $p_2 \leftarrow TournamentSelection(Pop,T)$\;
			\For{$M$ number of offspring}{
				Offspring $a \leftarrow p_{1}$\;
				\tcc{crossover}
				\If{Random(0,1) $<\mathcal{X}$} {
					Perform uniform crossover with $a$ and $p_2$\;
				}
				\tcc{mutation}
				\For{each parameter $x$ in $a$} {
					\If{Random(0,1) $<\mu$} {
						$x\leftarrow x + Random(-s,s)$\;
					}
				}         
				\tcc{evaluate offspring with surrogate model}
				$a.utility \leftarrow Rating(\mathcal{R},a,\mathcal{A})$\;
			}
			\tcc{select, evaluate, and add the most promising offspring}
			Evaluate the best utility offspring with the real objective function $k$ times\;
			Add offspring to $\mathcal{A}$\;
			$r \leftarrow NegativeTournamentSelection(Pop,T)$\;
			Replace $r$ with offspring\;
		}
    	\caption{Surrogate-assisted GA with pre-selection and static sampling}
    	\label{alg:sea}
    \end{algorithm}
\end{figure}
 
\begin{figure}[tbh]
    \begin{algorithm}[H]
    	\SetNoFillComment
    	\small
    	\DontPrintSemicolon
		{\bf Input:} fitted regression model $\mathcal{R}$, candidate $a$, evaluated archive $\mathcal{A}$\;
		{\bf Output:} model expected improvement of $a$\;
		$ei \leftarrow 0$\;
		$mean, std \leftarrow \mathcal{R}.predict(a)$\;
		\If{std != 0} {
			$imp \leftarrow BestFitness(\mathcal{A}) - mean$ \tcp*{minimising}
			$z \leftarrow imp / std$\;
			\tcc{$cdf()$ is the standard normal cumulative distribution function}
			\tcc{$pdf()$ is the standard normal probability density function}
			$ei \leftarrow imp \times cdf(z) + std \times pdf(z)$\;
		}
		return $ei$\;
    	\caption{Expected improvement rating function}
    	\label{alg:ei}
    \end{algorithm}
\end{figure}
 
\subsection{Results}

Since each simulation run is costly, we initially explored the case with $k=1$. However, after 400 evaluations/simulations, the performance of the fittest individual discovered by the GA was not significantly different than the fittest individual in the initial population, $p>0.05$ using a Wilcoxon rank-sums test, showing that the GA is severely misled by the significant variance in simulation runs with the same parameter set.

\begin{figure}[tbh]
	\centering 
	\includegraphics[width=\linewidth]{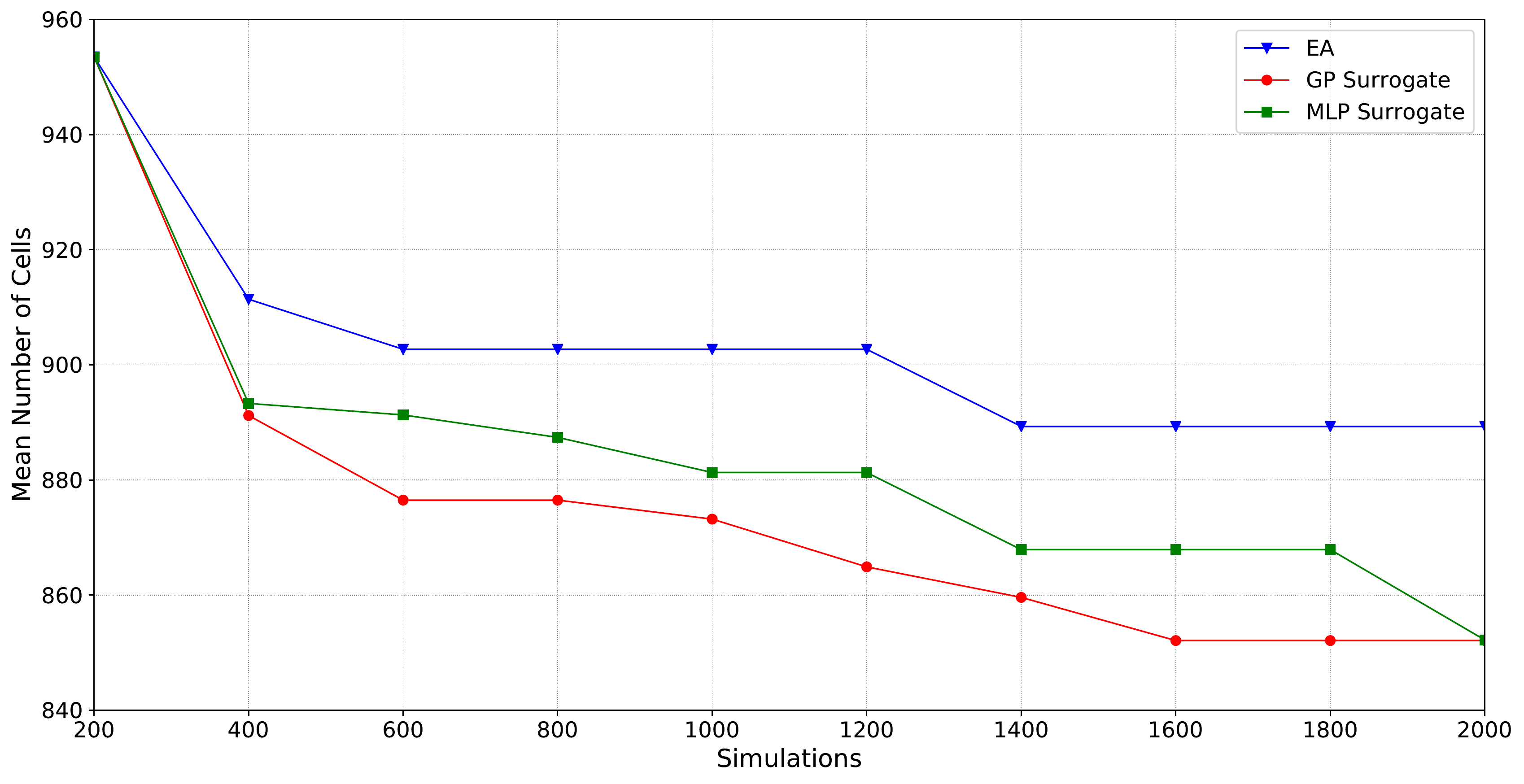}
	\caption{Fittest individuals on the PhysiCell anti-cancer biorobots simulator. GA (triangle), GP-assisted GA (circle), and MLP-assisted GA (square).}
	\label{fig:physicell_perf}
\end{figure}

Figure~\ref{fig:physicell_perf} shows the mean number of cells resulting from the fittest individual discovered for each algorithm with $k=10$. After evaluating 200 candidates (2000 simulations), the best GP-assisted solution (mean = 852.10, SD = 40.43, samples = 10, min = 773, median = 870, kurtosis = -0.89) and the best MLP-assisted solution (mean = 852.20, SD = 39.84, samples = 10, min = 791, median = 853, kurtosis = -0.74) are significantly less than the best GA solution without surrogate assistance (mean = 889.30, SD = 32.75, samples = 10, min = 831, median = 898, kurtosis = -0.83), $p\le0.05$ using a Wilcoxon rank-sums test. There is no significant difference between the surrogate models. All algorithms found solutions with a significantly lower mean number of cells than the best individual in the initial population (mean = 953.50, SD = 41.59, samples = 10, min = 887, median = 947, kurtosis = -1.10), $p\le0.05$.

\begin{figure}[tbh]
	\centering 
	\includegraphics[width=\linewidth]{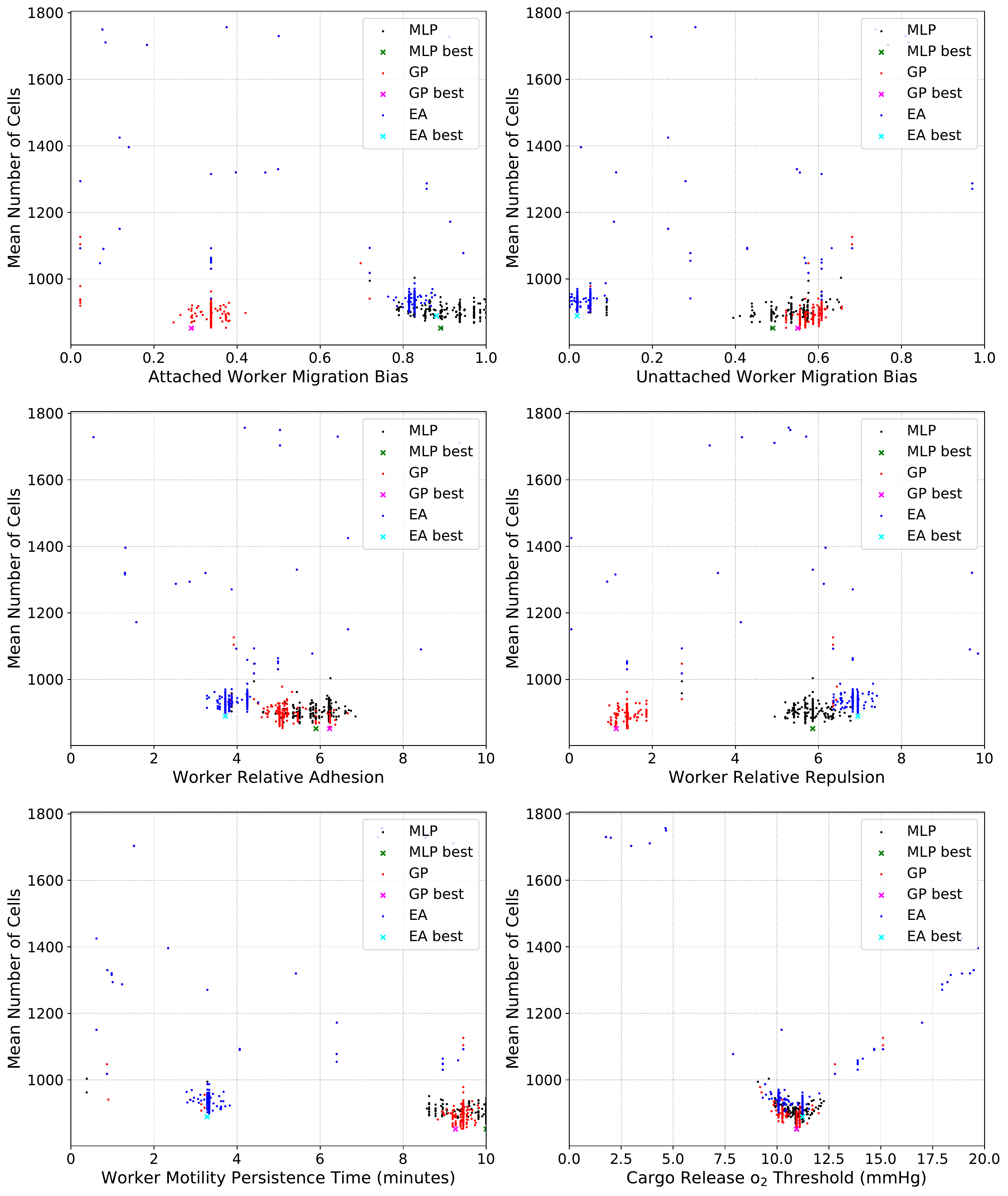}
	\caption{Scatter plot of all evaluated individuals on the PhysiCell anti-cancer biorobots simulator.}
	\label{fig:physicell_scatter}
\end{figure}

Figure~\ref{fig:physicell_scatter} shows scatter plots of all evaluated individuals for each evolved parameter. As can be seen, both surrogate-assisted EAs identified best solutions with a mean of 852 cells and these solutions have similar values for 4 of the parameters: $\approx0.5$ unattached worker migration bias; $\approx6$ worker relative adhesion; $\approx10$ worker motility persistence time; and $\approx11$ cargo release o$_2$ threshold. However, the MLP-assisted model achieved this with an attached worker migration bias of 0.89, whereas the GP-assisted solution was 0.29. Additionally, the worker relative repulsion was 5.9, compared with 1.13 for the GP-assisted model.

There appears to be a clear funnel with a minimum at $\approx11$ for the cargo release o$_2$ threshold, suggesting that this is the global optima for the parameter. \cite{Ghaffarizadeh:2018} used an initial cargo release o$_2$ threshold of 10, finding that ``once enough cancer cells were killed, hypoxia was reduced so that worker cells clustered near the oxygen minimum, but no longer released their cargo''. Increasing the threshold to 15 ``reduced but did not eliminate this behaviour''. The results of these simulations suggest that using a threshold of 11 results in the best performance. As \cite{Ghaffarizadeh:2018} note, ``the cargo release rules need to be carefully engineered. Such a system could potentially activate and deactivate to keep a tumour cell population in control, and to reduce hypoxia [which is known to drive cancer cell adaptation to more aggressive phenotypes~\citep{Wilson:2011,Eales:2016}]''.

\begin{figure}[tbh]
	\centering 
	\subfigure[Day 7]{\includegraphics[width=0.45\linewidth]{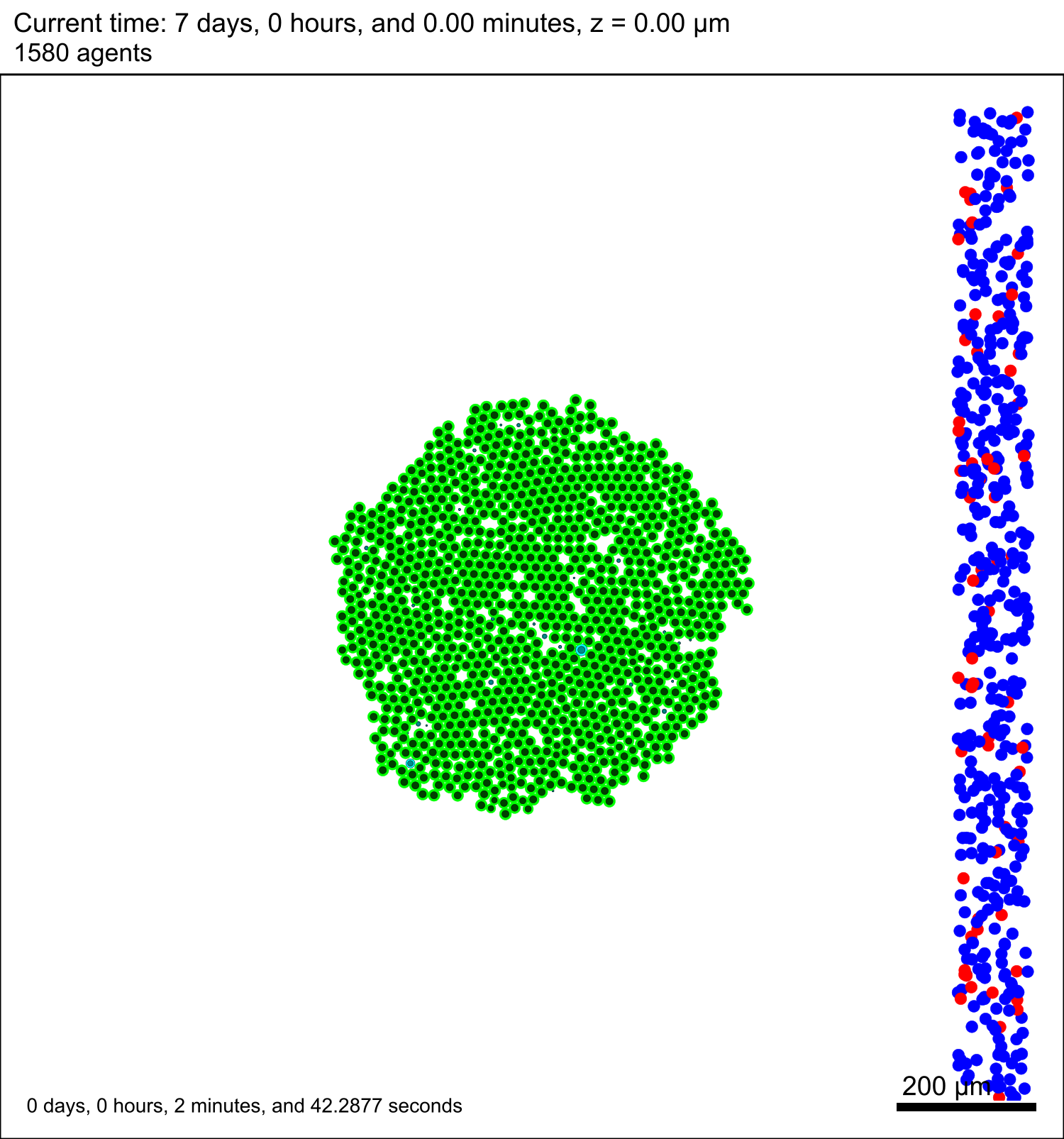}} 
	\subfigure[Day 8]{\includegraphics[width=0.45\linewidth]{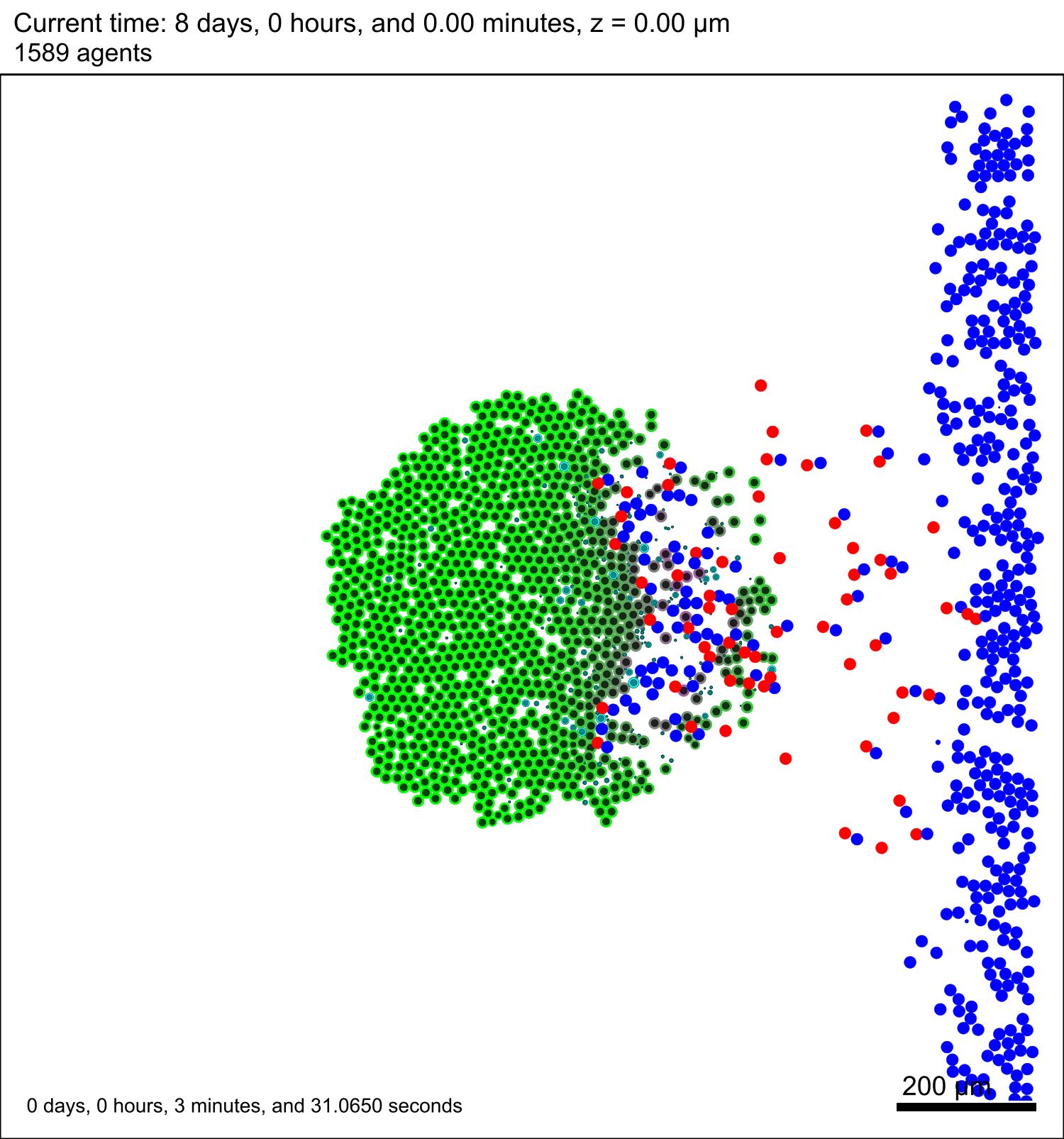}}\\
	\subfigure[Day 9]{\includegraphics[width=0.45\linewidth]{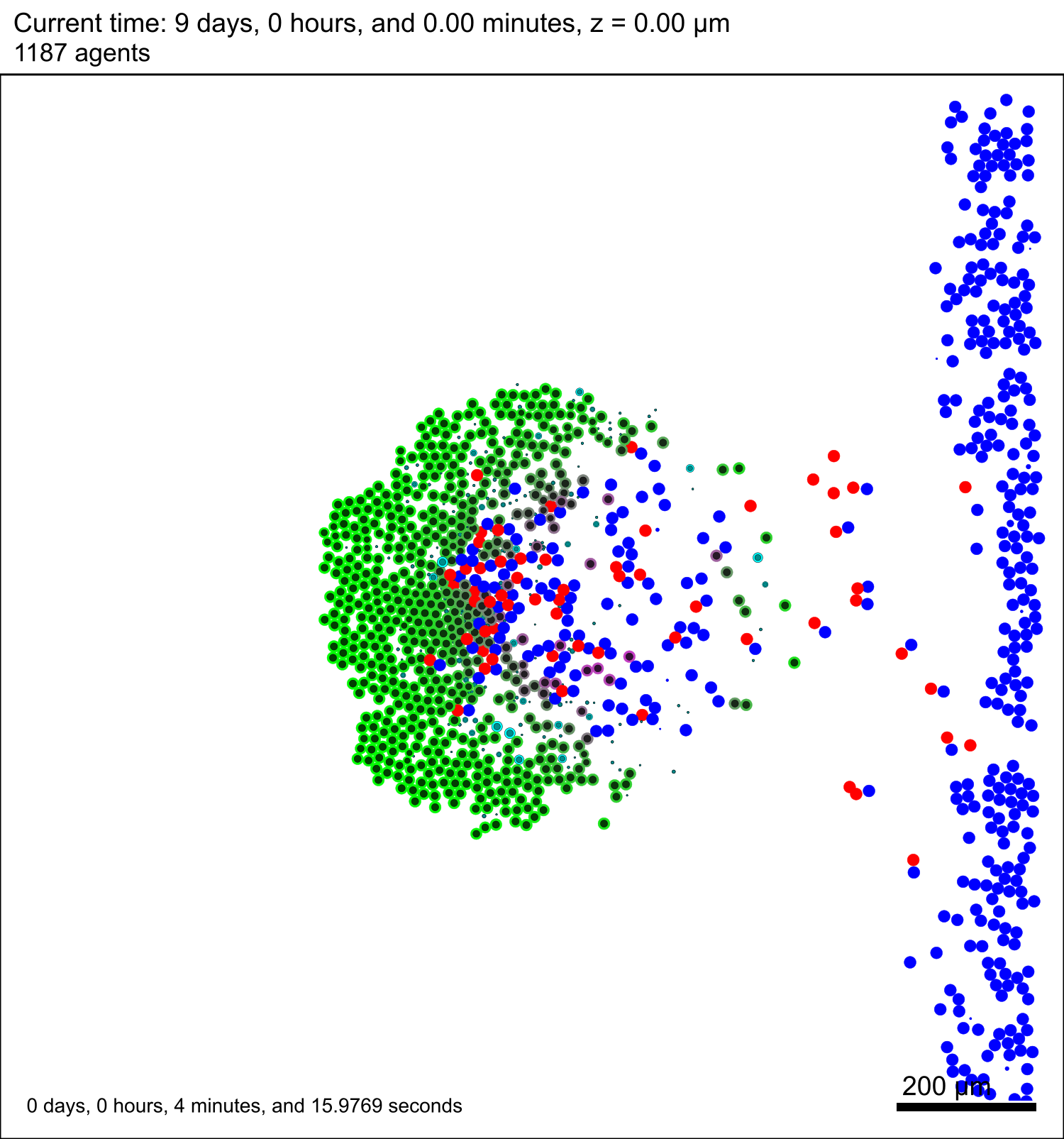}} 
	\subfigure[Day 10]{\includegraphics[width=0.45\linewidth]{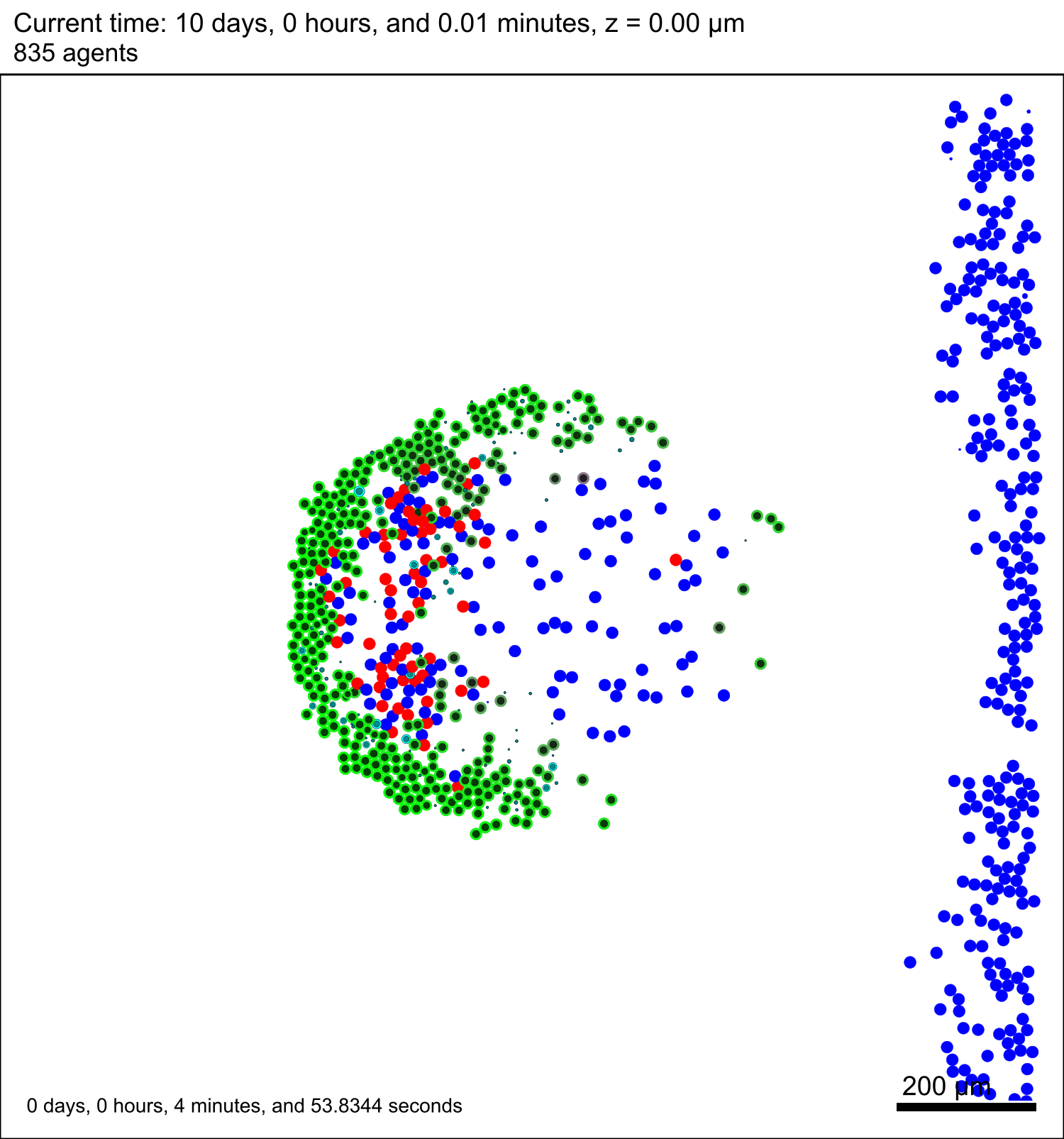}}\\
	\caption{Example run of the fittest evolved individual with the GP-surrogate model on the PhysiCell anti-cancer biorobots simulator after 200 candidate evaluations (2000 simulations.) Shown are the worker cells (red), cargo cells (blue), and tumour cells (green) after 7 days of tumour growth and each subsequent day of treatment. Attached worker migration bias = 0.29; unattached worker migration bias = 0.55; worker relative adhesion = 6.24; worker relative repulsion = 1.13; worker motility persistence time = 9.26; cargo release o$_2$ threshold = 10.94. Mean number of cells after 10 simulated days = 852.}
	\label{fig:physicell_final}
\end{figure}

An example run of the fittest evolved individual with the GP-surrogate model is shown in Figure~\ref{fig:physicell_final}, showing that the worker cells appear to disperse evenly and effectively deliver the cargo to the tumour.

\section{Conclusions}

This article has shown that EAs are able to effectively explore the parameter space of biophysical properties within the agent-based multicellular simulator, PhysiCell. EAs successfully minimised the number of cancerous cells after a period of simulated treatment. Both surrogate-assisted algorithms were found to outperform the standard GA, thereby reducing the number of expensive simulations required. No significant difference in the resulting number of cancerous cells was observed between the models, showing the robustness of the overall pre-selection with static sampling approach employed. By exploring both surrogate models that extrapolate and models that interpolate, we were able to identify different parameter sets that achieved similar reductions in cancerous cells. Thus, we have demonstrated the use of efficient EAs within a high-throughput computing approach to find therapeutic design optima that maximise tumour regression. 

\section*{Acknowledgement}

This work was supported by the European Research Council under the European Union's Horizon 2020 research and innovation programme under grant agreement No. 800983. 

%\section*{References}
%\bibliography{abrv,references}

\end{document}